# Design of Resistive Frequency Selective Surface based Radar Absorbing Structure – A Deep Learning Approach

Vijay Kumar Sutrakar
*Scientist, ADE, DRDO*
Bangalore, India
vks.ade@gov.in

Nikhil Mogre
*CE, ADE, DRDO*
Bangalore, India
nikhilmogre1998@gmail.com

Anjana P K
*CE, ADE, DRDO*
Bangalore, India
pkanjana505@gmail.com

Abhilash P V
*CE, ADE, DRDO*
Bangalore, India
abhilashpv33@gmail.com

*Abstract*—In this paper, deep learning-based approach for the design of radar absorbing structure using resistive frequency selective surface is proposed. In the present design, reflection coefficient is used as input of deep learning model and the Jerusalem cross based unit cell dimensions is predicted as outcome. Sequential neural network based deep learning model with adaptive moment estimation optimizer is used for designing multi frequency band absorbers. The model is used for designing radar absorber from L to Ka band depending on unit cell parameters and thickness. The outcome of deep learning model is further compared with full-wave simulation software and an excellent match is obtained. The proposed model can be used for the low-cost design of various radar absorbing structures using a single unit cell and thickness across the band of frequencies.

## I. INTRODUCTION

In recent years, there has been a noteworthy increase in the need for stealth vehicles in military applications. Radar cross-section (RCS) is commonly used to quantify a vehicle's stealth capability or low detectability. Advanced materials called radar-absorbing structures (RAS) are made especially to reduce RCS. Jaumann absorbers and the Salisbury screen are examples of early RAS technology used in stealth designs [1]. These designs were hampered by their substantial thickness and restricted bandwidth, which rendered them too bulky to be used in military aircraft. Later, frequency selective surface (FSS) based radar-absorbing structures came as promising alternative. These structures can be utilized artificially by tuning various resistivity/patterns for manipulating electromagnetic (EM) waves than previous designs [2]. Their two-dimensional, thin-layer geometry enables manufacturing to be simpler and has been widely used for multi-band/broadband RCS reduction applications [3]–[7]. Trial-and-error techniques, parametric optimization, and circuit-based analysis are the mainstays of the traditional FSS based design process [8]–[11]. It takes significant amount of computational time, specialist expertise, and extensive computational resources to utilize full-wave simulation software.

In recent past, machine learning (ML) has become as a popular alternative to conventional design technique in various branches of science and engineering including [12]–[15]. Deep learning (DL), which is a branch of ML, has been used to reduce the computational difficulties of traditional design approaches is the design of FSS based RAS [13]–[15]. Recent advancements have introduced ML/DL based models designed to predict the dimensions of unit cell for a specified reflection coefficient [16]. The first model, utilizing ML techniques, demonstrates exceptional accuracy in predicting the dimensions of the unit cell (represented as a scaling factor for given values of *a*, *b*, *c*, and *d*) for a fixed sample thickness. The second model, based on DL, excels in predicting both the scaling factor and the sample thickness [16]. Additionally, [16] highlights ongoing research aimed at extending the prediction of reflection coefficients to account for variations in unit cell size, shape, and operation across different frequency bands.

In this paper, geometrical dimensions of unit cell are predicted from a given reflection coefficient data utilizing deep learning techniques. Section II outlines the details of electromagnetic and DL modelling and simulation. Section III focuses on presenting the results and their interpretation, while Section IV offers concluding remarks along with suggestions for future research directions.

## II. MODELLING AND SIMULATION DETAILS

### A. Electromagnetic modelling and simulation details

In the present study, a unit cell (consist of three layers) featuring a Jerusalem cross pattern (with parameters *a*, *b*, *c*, *d*, and *t*), using a perfectly electrical conducting (PEC) material for the cross, with each arm having a resistance of 100 ohms is considered [16]. The first and third layers are made of FR4 material, which has a dielectric constant of $\epsilon_r = 4.4$ and a loss tangent of $\tan \delta = 0.02$. The second layer consists of the Jerusalem cross unit cell bonded to an FR4 substrate with a thickness of 0.125 mm.

Floquet port technique is used for generating the input of the deep learning model, i.e. reflection coefficient utilizing full-wave simulation technique [17]. Floquet port combined with periodic boundary conditions on the sides and a perfectly electrical conducting (PEC) boundary on the bottom layer is used. Extensive EM simulations were performed with parametric sweeps across the variables $a$, $b$, $c$, $d$, and $t$ to generate reflection coefficient data. Details are provided in Table I. The dimensions of $a$, $b$, $c$, and $d$ is chosen such that it does not cross the boundary of unit cell. Outer dimension of the unit cell, i.e. $a$ is varied from 3.5 to 7 with a step size of 0.5 mm. Further, for a given fixed value of $a$, the parameter $b$, $c$, $d$ is varied (refer Table I for further details). For example, for $a$ = 3.5 mm; $b$, $c$, $d$ is varied from 1.5 to 3.4, 0.25 to 1.5, 1.0 to 3.4, respectively. Similarly, for $a$ = 4.0 mm; $b$, $c$, $d$ is varied from 1.5 to 3.8, 0.25 to 1.4, 1.0 to 3.8, respectively. It is also to be noted in the geometric variation is that the variable $e$ is dependent on $a$ and $b$ which is given by,

$$e = (a-b)/2 \quad (1)$$

The data was collected over a frequency varies from 1 GHz to 30 GHz with a step size of 0.05 GHz.

In traditional design approaches, the dimension of Jerusalem cross, characterized by parameters $a$, $b$, $c$, $d$, and $t$ are input variables for electromagnetic simulations. The output of these simulation is the reflection coefficient or absorption data. Fig. 1 illustrates the schematic of a conventional EM simulation process. The ultimate objective of these simulations is to achieve the relevant reflection coefficient or absorption over a particular frequency range. Conventional unit cell based design can be complex and may result in longer design cycles. Therefore, an inverse approach is proposed where the dimensions of unit cell will be predicted (as output) from deep learning model using reflection coefficient (as input).

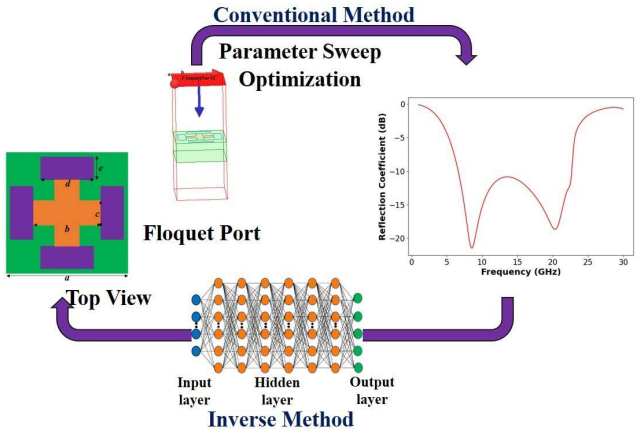

Fig. 1. Conventional electromagnetic design approach versus the inverse approach using deep learning

B. *Deep learning modelling and simulation details*

 *1) Case – 1 Prediction of unit cell dimensions for a given thickness:* In DL Model, i.e. sequential neural network [18],

TABLE I
PARAMETER SWEEP FOR DATA GENERATION

| $a$ (mm) | $b$ (mm) | $c$ (mm) | $d$ (mm) |
|---|---|---|---|
| 3.5 | 1.5 - 3.4 | 0.25 - 1.5 | 1.0 - 3.4 |
| 4.0 | 1.5 - 3.8 | 0.25 - 1.4 | 1.0 - 3.8 |
| 4.5 | 1.5 - 4.4 | 0.25 - 1.4 | 2.0 - 4.4 |
| 5.0 | 2.0 - 4.8 | 1.20 - 1.8 | 1.0 - 4.8 |
| 5.5 | 2.5 - 5.3 | 1.70 - 2.3 | 1.0 - 5.3 |
| 6.0 | 3.0 - 5.8 | 2.00 - 2.8 | 2.0 - 5.8 |
| 6.5 | 3.0 - 6.3 | 2.00 - 2.8 | 2.0 - 6.2 |
| 7.0 | 3.5 - 6.9 | 2.60 - 3.3 | 3.0 - 6.8 |

7600 samples are considered for a given thickness of 2.0 mm with varying unit cell parameters (refer Table I for further details). The DL model consist of Adaptive Moment Estimation (Adam) optimizer [19]. Further details are provided in Table II. The mean squared error (MSE) loss function [20] and R-Squared are used for evaluating the model performance. The Adam optimizer was chosen over other optimization algorithms for its efficiency and ability to adapt learning rates during training.

The dataset was split into 80:20 ratio for training and testing sets. The PCA technique was employed to transform the original input data into a lower-dimensional representation, resulting in 300 principal components that capture the most significant variance in the data [21]. This can help simplify the model and potentially improve its generalization performance. This reduction was carried out to alleviate the curse of dimensionality and improve model performance. L2 regularization is employed to address overfitting concerns [22]. The He Normal initializer [23] was used for weight initialization. It is known for its effectiveness in mitigating the vanishing gradient problem and enabling smoother training. Batch normalization was applied after each hidden layer [24]. It normalizes the inputs to hidden layers, leading to improved training speed and reduced sensitivity to weight initialization [25]. MSE versus number of Epochs for different number of hidden layers are shown in Fig. 2. Six hidden layers are used in the present DL model and it has loss (Error) for testing data is 0.06 and R-Squared is 0.95.

TABLE II
ARCHITECTURE OF DEEP LEARNING MODEL

| Layers | Dimensions | Normalization | Activation Function |
|---|---|---|---|
| Input | 300x1 | - | - |
| Hidden Layer 1 | 112x1 | Batch Normalization | Leaky ReLU |
| Hidden Layer 2 | 112x1 | Batch Normalization | Leaky ReLU |
| Hidden Layer 3 | 112x1 | Batch Normalization | Leaky ReLU |
| Hidden Layer 4 | 8x1 | Batch Normalization | Leaky ReLU |
| Hidden Layer 5 | 8x1 | Batch Normalization | Leaky ReLU |
| Hidden Layer 6 | 8x1 | Batch Normalization | Leaky ReLU |
| Output | 4x1 | - | - |

Subsequently, the same DL model has been tested for another RAS with thickness of 4.0 mm. The dataset contains approximately 8000 samples for RAS with thickness of 4.0

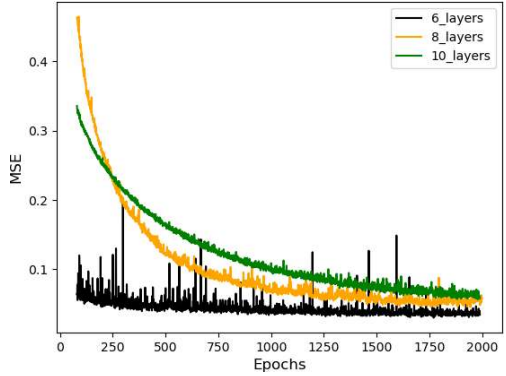

Fig. 2. MSE versus Epochs

mm, in which the variation is same as that of mentioned in Table I.

For further evaluation of the model, the same model architecture is used to build a new model for the prediction of unit cell parameters for a given thickness of 4.0 mm. The performance of new model is found excellent. Results are discussed in the subsequent section. It indicates that the architecture is sufficient enough to predict the geometric parameters of a unit cell for a given thickness.

*2) Case – 2 Prediction of a, b, c, d and thickness:* In the section Case - 1, a separate DL model is developed for a given RAS thickness. In this section, a new DL model combining the dataset of 2.0 mm and 4.0 mm thickness with the same architecture is developed. The performance of the new DL model is found excellent. In case - 2, the geometric parameters as well as the thickness of the unit cell are predicted as output. Further new dataset with thickness varying from 1 to 10 mm has been added. In this extended model, there are about 76000 samples considered. The architecture used for training the model with the extended dataset is the same which is mentioned in Table I for thickness 1 to 10 mm.

### III. RESULTS AND DISCUSSIONS

#### A. Case – 1 Prediction of a, b, c, d for a given thickness

Firstly, RAS with 2.0 mm thickness is studied. Four random samples (reflection coefficient data) have been taken from the test data (taken from samples of test data of 2.0 mm RAS, refer Section II (B), Case – 1) for evaluating the unit cell performance. The predicted values of *a, b, c, d* for a given RAS with 2.0 mm thickness and the respective error in percentage is shown in Table III. Subsequently, those unit cell parameters are used in EM simulations (refer Section II (A) for further details) for the generating the reflection coefficients. The comparison of true and predicted reflection coefficient performance of four random RAS samples with 2.0 mm thickness are shown in Fig. 3. The proposed DL model has loss (Error) of 0.06 and R-Squared of 0.95 for testing data and it is able to predict the RAS dimensions accurately. It can be seen from the four random samples, the maximum errors for predicting the dimensions are 10% (except for the sample (c)). However, there is very close match in the true and predicted values of reflection coefficient across the frequency band for sample c (refer Fig 3(c) for further details). Results indicate that there could be multiple solutions for the same reflection coefficient and DL model is able to predict it correctly. It can also be noted that 2.0 mm RAS is able to give reflection coefficient less than ~ 10 dB from 20 GHz to 30 GHz frequencies.

Subsequently, four random samples have been taken from the test data of RAS with 4.0 mm of thickness (refer Section II (B), case – 1). The predicted values of *a, b, c, d* for a given RAS with 4.0 mm thickness and the error in percentage is shown in Table IV. Subsequently, those unit cell parameters are used in EM simulations (refer Section II (A) for further details) for the generating the reflection coefficients. The comparison of true and predicted reflection coefficient performance of four random RAS samples with 4.0 mm thickness are shown in Fig. 4. It can be seen from the four random samples, the maximum errors for predicting the dimensions are 10% (except for the few unit cell parameters of samples (b) and (c)). However, it can be seen in the Fig. 5(b) and 5(c) that there is very close match in the true and predicted values of reflection coefficient across the frequency band. As discussed previously, results indicate that there could be multiple solutions for the same reflection coefficient and DL model is able to predict those unit cell parameters correctly. Also, it can be seen that simulated reflection matches well with the true reflection coefficient for frequency varying from 1 GHz to 30 GHz. It can also be seen that 4.0 mm RAS is able to give reflection coefficient less than ~ 10 dB from 8 GHz to 27 GHz frequencies.

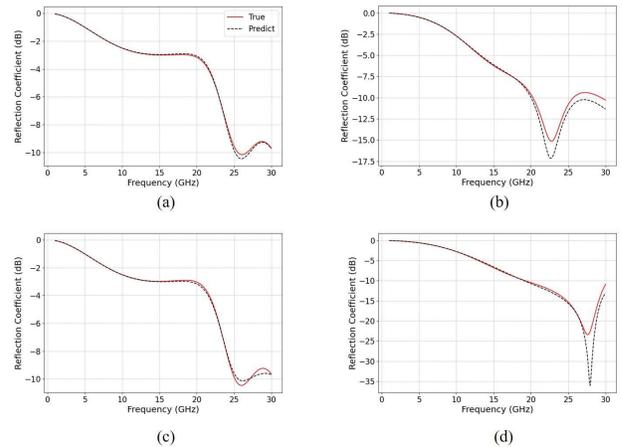

Fig. 3. Reflection coefficient prediction for RAS with 2 mm thickness

#### B. Case – 2 Prediction of a, b, c, d and thickness

A DL model is trained with thickness from 1 to 10 mm (refer section II (B) (2) for further details) with approximately 76000 samples, in which data is considered as 80:20 ratio for training and testing. The architecture used for training the model with the extended dataset is the same which is mentioned in Table I for thickness 1 to 10 mm.

TABLE III
TRUE, PREDICTED AND PERCENTAGE ERROR OF UNITCELL VALUE OF RAS WITH 2.0 MM THICKNESS

| Sl No | True Value (mm) | | | | Prediction Value (mm) | | | | Percentage Error (%) | | | |
|---|---|---|---|---|---|---|---|---|---|---|---|---|
| | a | b | c | d | a | b | c | d | a | b | c | d |
| (a) | 6.88 | 3.92 | 2.83 | 5.46 | 6.89 | 4.01 | 2.84 | 5.36 | 0.15 | 2.30 | 0.35 | 1.83 |
| (b) | 7.89 | 2.89 | 1.71 | 3.17 | 8.12 | 2.89 | 1.60 | 3.45 | 2.92 | 0.00 | 6.43 | 8.83 |
| (c) | 5.11 | 2.45 | 1.44 | 1.94 | 5.31 | 2.16 | 1.08 | 2.59 | 3.91 | 11.84 | 25.00 | 33.51 |
| (d) | 6.90 | 4.07 | 2.75 | 5.37 | 6.88 | 3.92 | 2.83 | 5.46 | 0.29 | 3.69 | 2.91 | 1.68 |

TABLE IV
TRUE, PREDICTED AND PERCENTAGE ERROR OF UNITCELL VALUE OF RAS WITH 4.0 MM THICKNESS

| Sl No | True Value (mm) | | | | Prediction Value (mm) | | | | Percentage Error (%) | | | |
|---|---|---|---|---|---|---|---|---|---|---|---|---|
| | a | b | c | d | a | b | c | d | a | b | c | d |
| (a) | 6.0 | 4.5 | 2.25 | 4.25 | 6.07 | 4.56 | 2.22 | 4.19 | 1.17 | 1.33 | 1.33 | 4.41 |
| (b) | 3.5 | 3.0 | 1.5 | 1.5 | 3.52 | 2.85 | 1.41 | 1.78 | 0.57 | 5.00 | 6.00 | 18.67 |
| (c) | 5.5 | 4.25 | 2.3 | 4.25 | 5.5 | 4.18 | 2.2 | 4.65 | 0.00 | 1.65 | 4.35 | 9.41 |
| (d) | 3.5 | 3.0 | 0.5 | 2.0 | 3.54 | 3.0 | 0.61 | 2.0 | 1.14 | 0.00 | 22.00 | 0.00 |

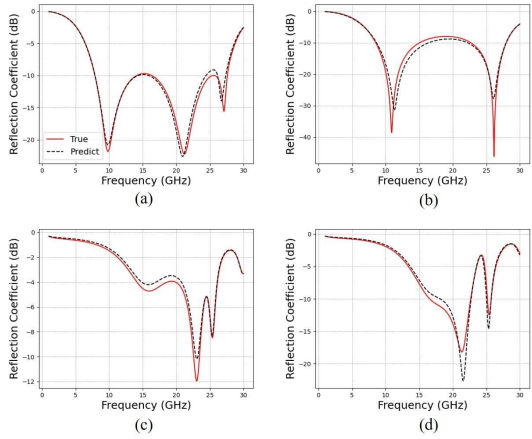

Fig. 4. Reflection coefficient prediction for RAS with 4 mm thickness

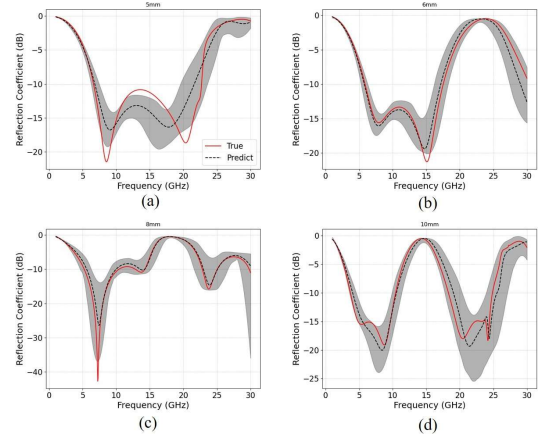

Fig. 5. Reflection coefficient prediction for various thickness from test data

For evaluation, samples from different thicknesses are considered (refer Table V for further details). The true and predicted values as well as the respective error in percentage is shown in Table V and the reflection coefficients are shown in Fig. 5. Further, ±5% variation in true values of reflection coefficient are also shown as shaded region in Fig. 5. Result indicates an excellent match between the true and predicted reflection coefficient.

Subsequently, four random samples have been taken from the data base of RAS with varying thickness (refer Section II (B), Case – 2). True and predicted reflection coefficient of RAS with various thickness with corresponding percentage error in Table VI and Fig 6. The proposed DL model predict the reflection coefficient accurately across the frequency bands (varying from 1 GHz to 30 GHz). Result also indicates that the DL model could be utilized for broad band RAS design.

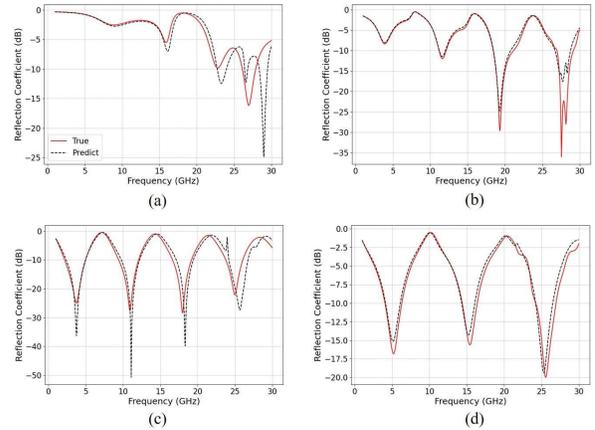

Fig. 6. True and predicted curves for random samples

TABLE V
True, predicted and percentage error for different thickness (case-2)

| Sl No | True Value (mm) | | | | | Prediction Value (mm) | | | | | Percentage Error (%) | | | | |
|---|---|---|---|---|---|---|---|---|---|---|---|---|---|---|---|
| | a | b | c | d | t | a | b | c | d | t | a | b | c | d | t |
| (a) | 6.5 | 3.0 | 2.0 | 2.0 | 5.0 | 6.08 | 3.1 | 2.0 | 2.03 | 5.05 | 6.4 | 3.3 | 0.0 | 1.5 | 1.0 |
| (b) | 3.5 | 1.5 | 1.25 | 1.5 | 6.0 | 4.16 | 1.63 | 0.76 | 2.02 | 6.17 | 18.0 | 8.6 | 36.8 | 34.6 | 2.8 |
| (c) | 4.5 | 3.5 | 0.25 | 2.5 | 8.0 | 4.01 | 3.15 | 0.25 | 1.99 | 8.0 | 10.8 | 10.0 | 0.0 | 20.4 | 0.0 |
| (d) | 6.0 | 3.25 | 2.25 | 2.5 | 10.0 | 5.97 | 3.14 | 2.14 | 2.55 | 9.71 | 0.5 | 3.3 | 4.8 | 2.0 | 2.9 |

TABLE VI
True values, prediction values, and percentage errors

| Sl No | True Value (mm) | | | | | Prediction Value (mm) | | | | | Percentage Error (%) | | | | |
|---|---|---|---|---|---|---|---|---|---|---|---|---|---|---|---|
| | a | b | c | d | t | a | b | c | d | t | a | b | c | d | t |
| (a) | 5.5 | 5.3 | 2.20 | 3.75 | 4.0 | 5.45 | 5.25 | 1.99 | 3.55 | 3.90 | 0.9 | 0.9 | 9.0 | 5.0 | 2.5 |
| (b) | 7.0 | 3.75 | 2.85 | 5.50 | 7.0 | 7.05 | 3.87 | 2.93 | 5.86 | 7.09 | 0.7 | 3.0 | 2.0 | 6.0 | 1.0 |
| (c) | 5.5 | 4.75 | 1.70 | 4.50 | 9.0 | 5.37 | 4.62 | 1.77 | 4.36 | 9.06 | 2.0 | 2.0 | 4.0 | 3.0 | 0.6 |
| (d) | 6.5 | 3.0 | 2.25 | 3.75 | 10.0 | 6.22 | 3.09 | 2.4 | 3.66 | 9.80 | 4.0 | 3.0 | 6.0 | 2.0 | 2.0 |

IV. CONCLUSION

In this paper, radar absorbing structure design using deep learning model is demonstrated. The study applies deep learning to predict geometric parameters of unit cell using their electromagnetic properties; hence, enabling an efficient design and relative ease compared to classical approaches for creating radar absorbing structures. This methodology has potential to save time and computational resources while maintaining high accuracy in the creation of radar-absorbing materials. The models explained in this paper are able to predict the unit cell dimensions for a given fixed thickness as well as for different thickness. The predicted values of the models show excellent match with true values from L to Ka frequency bands.

ACKNOWLEDGEMENT

Authors would like to thank Shri. Y Dilip, Director, Aeronautical Development Establishment, Mr. Manjunath S M, Technology Director and Mr. Diptiman Biswas, Group Director for their support during the research work carried out at ADE, DRDO.